\documentclass[10pt,twocolumn,letterpaper]{article}

\usepackage{iccv}
\usepackage{times}
\usepackage{epsfig}
\usepackage{graphicx}
\usepackage{subfigure}
\usepackage{amsmath}
\usepackage{amssymb}
\usepackage{bm}
\usepackage{booktabs}
\usepackage{textcomp}
\usepackage{multirow}

\usepackage[pagebackref=true,breaklinks=true,letterpaper=true,colorlinks,bookmarks=false]{hyperref}

\iccvfinalcopy 

\def\iccvPaperID{2032} 

\ificcvfinal\fi
\def\A{{\mathbf A}}
\def\X{{\mathbf X}}
\def\W{{\mathbf W}}
\def\T{{\mathbf T}}
\def\x{{\mathbf x}}
\begin{document}

\title{Sequence Level Semantics Aggregation for Video Object Detection}

\author{
Haiping Wu$^{1}$\quad Yuntao Chen$^{3,4}$\quad Naiyan Wang$^{2}$ \quad Zhaoxiang Zhang$^{3,4,5}$ \\ 
$^{1}$ McGill University \qquad $^{2}$ TuSimple\\ 
$^{3}$ University of Chinese Academy of Sciences \\ 
$^{4}$ Center for Research on Intelligent Perception and Computing, CASIA\\ 
$^{5}$Center for Excellence in Brain Science and Intelligence Technology, CAS\\ 
{\tt\small haiping.wu2@mail.mcgill.ca \{chenyuntao2016, zhaoxiang.zhang\}@ia.ac.cn winsty@gmail.com}
}

\maketitle
\ificcvfinal\thispagestyle{empty}\fi

\begin{abstract}
Video objection detection (VID) has been a rising research direction in recent years. 
A central issue of VID is the appearance degradation of video frames caused by fast motion. 
This problem is essentially ill-posed for a single frame. 
Therefore, aggregating features from other frames becomes a natural choice. 
Existing methods rely heavily on optical flow or recurrent neural networks for feature aggregation. 
However, these methods emphasize more on the temporally nearby frames. 
In this work, we argue that aggregating features in the full-sequence level will lead to more discriminative and robust features for video object detection. 
To achieve this goal, we devise a novel Sequence Level Semantics Aggregation (SELSA) module. 
We further demonstrate the close relationship between the proposed method and the classic spectral clustering method, providing a novel view for understanding the VID problem. 
We test the proposed method on the ImageNet VID and the EPIC KITCHENS dataset and achieve new state-of-the-art results. Our method does not need complicated post-processing methods such as Seq-NMS or Tubelet rescoring, which keeps the pipeline simple and clean. 
\end{abstract}

\section{Introduction}
Recent years have witnessed fast progress in object detection using deep convolutional networks. 
Renewed detection paradigms~\cite{girshick2015fast,ren2015faster,he2017mask}, strong backbone~\cite{he2016deep,xie2017aggregated} and large scale datasets~\cite{lin2014microsoft,OpenImages} jointly push forward the limit of object detection. 

Video Object Detection (VID) has now emerged as a new challenge beyond object detection in still images. 
Thanks to the fast progress in still image object detection, detectors' performance on slow-moving objects in video object detection has somewhat saturated~\cite{zhu17fgfa}.
The main challenge now lies in the scenario where objects or cameras are under fast motion.

\begin{figure}[!t]
	\centering
	\includegraphics[width=0.5\textwidth]{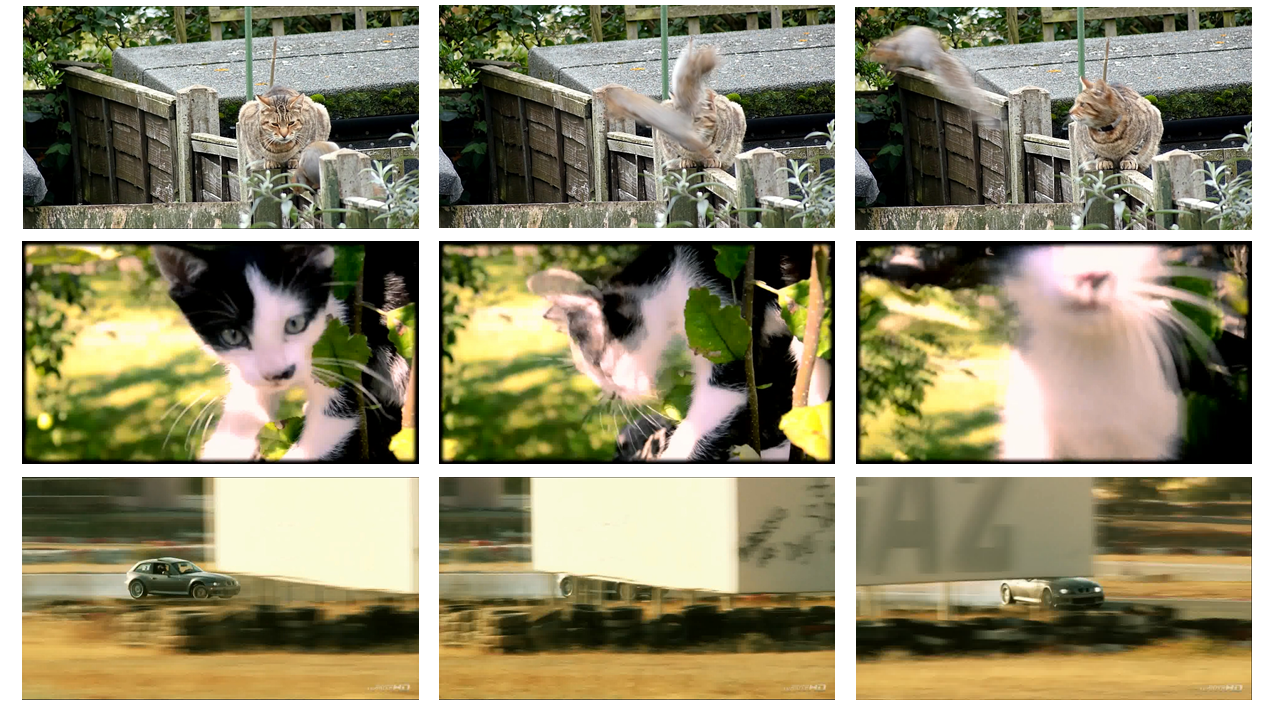}
	\caption{Challenges in video object detection. Motion blur, camera defocus and pose variation.}
	\label{fig:problem}
\end{figure}

Fast motion brings up image degradation unseen in the still image setting like motion blur, camera defocus and large pose variation as shown in Figure~\ref{fig:problem}.
Still image detectors often fail in these cases. 
On the other hand, a video provides far richer visual information than a still image. 
When the appearance of an object deteriorates in a frame, it is natural to include information from the video (e.g. nearby frames) to mitigate this degradation. 
The second and third columns in Figure~\ref{fig:problem} show various difficult sequences in VID. 
Though in these hard cases, there are still some frames more salient than the others. 
A good video object detector should be able to identify the salient views to refine its beliefs on those degraded views if they are (semantically) similar, either to support its beliefs or deny them. 
Note that useful information is not necessarily from temporal nearby frames, any objects share high similarity with the object of interest in any frames (even within the same frame) could contribute.

Post-processing methods try to incorporate video-level information by designing sophisticated rule set for linking bounding boxes generated by still image detectors. 
These two-stage methods are not jointly optimized and may lead to sub-optimal results. 
Instead, end-to-end feature aggregation utilizes motion information estimated from optical flow~\cite{zhu2017deep} or instance tracking~\cite{wang2018fully} for object feature calibration. 
Feature calibration methods heavily rely on accurate motion estimation, which is somewhat contradictory. 
In the circumstance of fast motion, the appearance of objects degrades drastically. 
Thus the results of optical flow are usually unsatisfactory in such cases, which makes it less helpful for VID task.

To lift this limitation in a principled way, we need to take a deeper look at the video itself. 
Existing works generally take video as sequential frames, and thus mainly utilize the temporal information to enhance the performance of a detector. 
For example, Flow Guided Feature Aggregation (FGFA)~\cite{zhu17fgfa} uses at most 21 frames during training and testing, which is less than 5\% of average video length. 
Instead of taking a consecutive viewpoint, we propose to treat video as a bag of unordered frames and try to learn an invariant representation of each class on the full sequence level. 
\emph{This reinterprets video object detection from a sequential detection task to a multi-shot detection task.}

In the multi-shot view, a video consists of clusters of objects, with each cluster containing hundreds even thousands of shots. 
The appearance degradation of an object is the manifestation of large intra-class feature variance. 
Thus reducing the feature variance lies in the core of addressing appearance changes. 
As mentioned before, temporal feature aggregation is a well-established way for feature variance reduction. However, it fails to utilize the rich information beyond a fixed time window. 

We take a further step by clustering and enhancing features in the entire sequence level. 
In this work, we present Sequence Level Semantics Aggregation (SELSA) method. 
We introduce SELSA module which is inspired by spectral clustering. 
Features of Region of Interests (ROI) are extracted from frames sampled from the whole video, and then go through our clustering module and transformation module. 
The enhanced features are handed to the detection head to get final detection results. 
Our method is thoroughly tested on the large scale ImageNet VID and EPIC KITCHEN datasets. 
We also design ablation experiments to demonstrate the effectiveness of proposed methods. 
We achieve 82.7 mAP with Faster-RCNN detector and ResNet-101 backbone and 84.3 mAP with ResNeXt-101 backbone, improving the state-of-the-art results by a large margin.
Additional experiments on EPIC KITCHENS~\cite{damen2018scaling} dataset show that our method generalize to more complex scenes.

In summary, our contributions are three folds:
\begin{enumerate}
	\item We first treat video detection as a sequence level multi-shot detection problem and then introduce a global clustering viewpoint of VID task for the first time.
	\item To incorporate such view into current deep object detection pipeline, we introduce a simple but effective Sequence Level Semantics Aggregation (SELSA) module to fully utilize video information.
	\item We test our proposed method on the large scale ImageNet VID and EPIC KITCHEN datasets and demonstrate significant improvement over previous methods.
\end{enumerate}

\section{Related Work}
In this section, we briefly review several works that are closely related to our method.

\subsection{Object Detection in Still Images}
Thanks to the success of deep neural networks, state-of-the-art detection systems~\cite{ren2015faster, dai2016r} are based on deep convolution neural networks (CNNs).
The typical two-stage detector R-CNN~\cite{girshick2014rich} first extracts regional features from backbone networks based on deep CNNs, and then classifies and refines the corresponding bounding boxes.
Fast R-CNN~\cite{girshick2015fast} proposed RoIPooling operation to speed up the regional feature extraction process.
Traditionally, region proposals are generated through selective search~\cite{uijlings2013selective}. 
The Regional Proposal Network (RPN) was proposed in Faster R-CNN~\cite{ren2015faster} to generate region proposals through deep CNNs, using backbone networks shared with Fast R-CNN. 
R-FCN~\cite{dai2016r} introduced position-sensitive RoIPooling operation, improving the detection efficiency by sharing the computation of regional features.

On the other hand, one-stage object detector directly predicts the bounding box of interest based on the extracted feature map from CNN. 
Without the extra stage, one-stage detector is usually faster than the two-stage counterpart. 
Representative works include YOLO~\cite{redmon2016you} and its variants~\cite{redmon2017yolo9000,redmon2018yolov3}, SSD~\cite{liu2016ssd} and its variants~\cite{fu2017dssd,lin2018focal}. 
Nevertheless, one-stage detector can hardly extend to more complicated tasks such as key point detection and instance segmentation. 
Similarly in our work, it can hardly be extended to extract proposal-level object semantic features. 
Thus we choose Faster R-CNN as our basic still image detector.

Recently, high-level relations among objects in object detection has been studied in~\cite{hu2018relation,wang2018non}. 
These works model the appearance and geometry relations among object proposals within a single image. 
This enables joint reasoning of objects and improves the accuracy. It could also be used as a duplicate removal step instead of NMS since the geometry relations is embedded. 
Similarly, our work also captures relations among objects.
However, we especially capture the relation measured by semantic similarity (objects of the same class across the video) instead of high-level interaction between objects (e.g person v.s glove in~\cite{hu2018relation}). 
We use these similarities to guide our feature aggregation and alleviate problems introduced by videos (fast motion).

\subsection{Object Detection in Videos}
For object detection in videos, the main challenge lies in how to utilize the rich information of videos (e.g. temporal continuity) to improve the accuracy as well as the speed upon still image detectors. 

Several previous works devised various post-processing techniques applied to the results of still image detectors by leveraging temporal information: 
Kang \etal \cite{kang2016object, kang2017t} proposed to suppress false positive detections via multi-context suppression (MCS) and propagate predicted bounding boxes across frames using the motion calculated by optical flow. Then a temporal convolution neural network is trained to  rescore the tubelets generated using visual tracking.
Feichtenhofer \etal \cite{feichtenhofer2017detect} performed single-frame object detection and object movements regression across frames (tracking) in a multi-task fashion. Then it links the detections across frames to object tubelets using the predicted movements, and re-weights detection scores in tubelets. 
Han \etal \cite{han2016seq} proposed Seq-NMS to form high score linkages using bounding box IoU across frames and then rescore the boxes associated with each linkage to the average or maximum scores of the linkage. 
These methods perform box-level post-processing upon still image detections, which could be sub-optimal since they are not optimized jointly. 
In contrast, our method manages to leverage video-level information at proposal-level by end-to-end optimization without post-processing steps.

Another line of work~\cite{kang2017t} focuses on utilizing optical flow to extract motion information to facilitate object detection.
However, such pre-computed optical flow is neither efficient nor task related.
Deep Feature Flow (DFF)~\cite{zhu2017deep} is the first work that adopts in-network fine-tuned optical flow computation. 
It utilizes the optical flow generated by FlowNet~\cite{dosovitskiy2015flownet} to propagate and align the features of selected keyframes to nearby non-keyframes, thus reducing redundant calculation and speeding up the system. 
FGFA~\cite{zhu17fgfa} is built on DFF~\cite{zhu2017deep}. However, its objective is to improve the accuracy by aligning and aggregating features from keyframes using optical flow. 
Based on DFF and FGFA, MANet~\cite{wang2018fully} adds an instance-level feature calibration and aggregation module besides the pixel-level one in FGFA, and then it combines these two levels through a motion pattern reasoning module. 
Furthermore, \cite{zhu2018towards} and \cite{STLattice2018CVPR} design more advanced feature propagation and keyframe selection mechanisms to improve the accuracy as well as the speed. 

Using optical flow to calibrate features across frames could be error-prone since object location, appearance and pose could change dramatically, where optical flow estimation becomes unreliable. 
Unlike these methods, our method does not intend to align features across frames by temporal information. 
We aggregate features on the proposal level, which makes our method more robust and superior.

Tripathi \etal \cite{tripathi2016context} trained a recurrent neural network to refine its initial detection results.
Lu \etal~\cite{lu2017online} used association LSTM to address the object association between consecutive frames. 
STMN~\cite{xiao2018video} used a Spatial-Temporal Memory module as the recurrent operation to pass the information through a video.
Unlike~\cite{xiao2018video}, our method does not need to pass information using memory modules in temporal order. We form clusters and aggregate features in a multi-shot view to capture the rich information of videos instead. 
Also, our clustering and feature aggregation are performed on instance-level features, where redundant pixel-level calculation is unnecessary. Moreover, it focuses more on subjects of interest.

\section{Method}
In this section, we first describe the motivation of our Sequence Level Semantics Aggregation (SELSA) method in Sec.~\ref{sec:motiv}. 
We then elaborate the details of our SELSA module in Sec.~\ref{sec:module}. 
We further interpret our method from the clustering view in Sec.~\ref{sec:spec}. 
Finally, we discuss the relation between our method and existing works in Sec.\ref{sec:discuss}.

\begin{figure*}[t]
	\centering
	\includegraphics[width=0.9\linewidth]{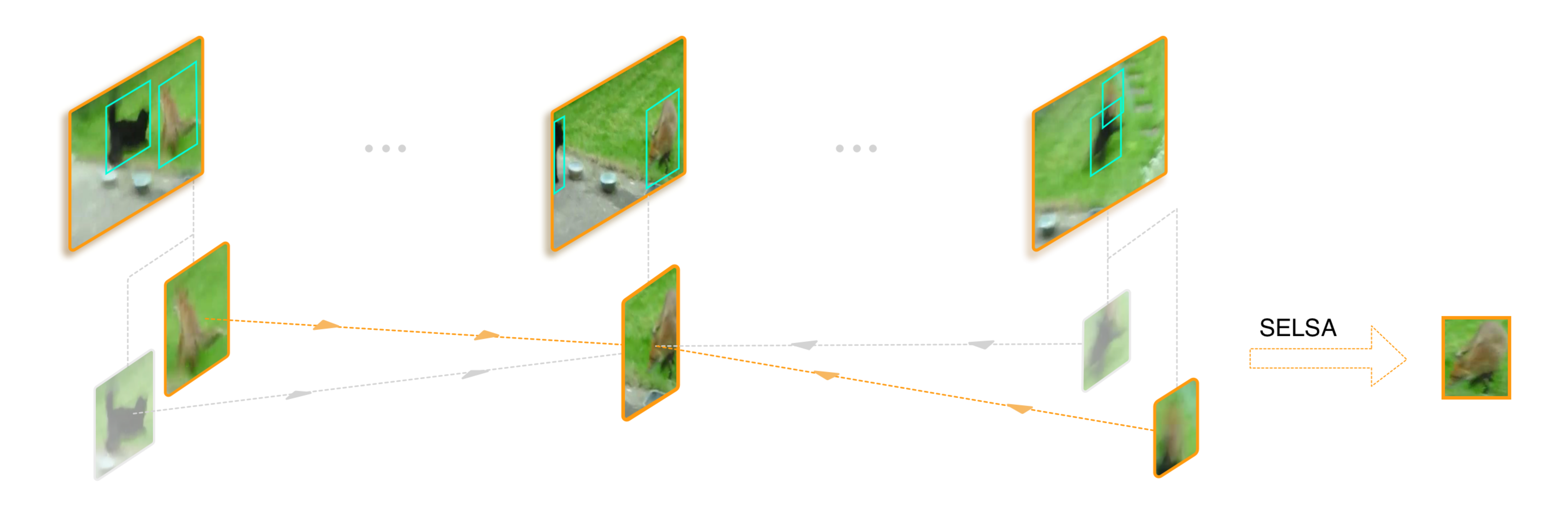}
	\caption{The overall architecture of the proposed model. We first extract proposals in different frames from the video, then the semantic similarities of proposals are computed across frames. At last, we aggregate the features from other proposals based on these similarities to obtain a more discriminative and robust features for object detection.}
	\label{fig:pipeline}
\end{figure*}

\subsection{Motivation}
\label{sec:motiv}
Feature aggregation is an effective way to mitigate the appearance degradation in video detection. 
The vital part of this method is to choose proper features for aggregation. 
Previous methods ~\cite{wang2018fully,zhu17fgfa} generally utilize features from a short temporal window. 
But appearance deterioration could span a wide time window and thus makes temporal-based methods less effective. 
Moreover, the frames may be highly redundant in a short time window and consequently weaken the advantage of feature aggregation. 
To address this problem, we propose to aggregate feature from the semantic neighborhood, which is not susceptible to the appearance degradation lasting in time.

\subsection{Sequence Level Semantics Aggregation}
\label{sec:module}
The ideal way for feature aggregation is to aggregate within the ground truth tracklet. 
But the golden association for proposals across frames is not available during test phase. 
Inspired by the ReID-based association which is popular in multi-object tracking system~\cite{wojke2017simple}, we propose to link proposals across space-time with their semantic similarities. 
This semantic feature based association approach is well known for its robustness to appearance change. 

\paragraph{Semantic Guidance} 
For each frame $f$, let $\X^f = \{\x_1^f, \x_2^f, \cdots\}$ be the proposals generated by the RPN network from Faster-RCNN. 
For a specific pair of proposals $(\x_i^k, \x_j^l)$, we measure the semantic similarity between them with the generalized cosine similarity:
\begin{equation}
\label{eq:similar}
w_{ij}^{kl} = \phi(\x_i^k)^T \psi(\x_j^l), 
\end{equation}
where $\phi(\cdot)$ and $\psi(\cdot)$ are some general transformation functions. 
Higher similarity indicates a higher chance of proposals being in the same category.\\

\paragraph{Feature Aggregation} 
After defining the similarity between proposals, the semantic similarity now serves as guidence for the reference proposal to aggregate features from other proposals. 
By aggregating across multiple proposals, the new proposal feature contains much richer information and should be robust against appearance variation like pose change, motion blur, and object deformation. 
Moreover, since the similarities are built on the proposal level, they are more robust compared with the optical flow which is computed on each position in feature maps.

In order to preserve the magnitude of the features after aggregation, we normalize the similarities with softmax function across all proposals. 
Formally, suppose that we are aggregating from randomly picked $F$ frames in the video with $N$ proposals produced in each frame, the aggregated feature for reference proposal is defined as:
\begin{equation}
\bar{\x}_i^k = \sum_{l \in \Omega}\sum_{j=1}^N w_{ij}^{kl} \x_j^l,
\end{equation}
where $\Omega$ is the set of frame indexes randomly selected for the aggregation. 
The SELSA module is fully differentiable and can be optimized end-to-end with standard SGD. 
After the aggregation, the enhanced proposal features are further fed into the detection header network for classification and bounding box regression. 
Figure~\ref{fig:pipeline} shows how the proposed SELSA module work.

\subsection{A Spectral Clustering Viewpoint}
\label{sec:spec}
Besides the simple and intuitive formulation of our method, we further reveal its close connection with the classic spectral clustering algorithm. This sheds light on how SELSA work from an intra-class variance reduction viewpoint.

With proposals $\X$ as nodes and similarity $\W$ as edges, we can define a semantic similarity graph $G = (\X, \W)$ on the proposals. 
From a probabilistic viewpoint, the random walk on graph $G$ is controlled by the stochastic matrix $\T$ which is obtained by normalizing each row in $W$ to sum 1. 
$\T_{ij}$ describes the transition probability from proposal $i$ to proposal $j$ during a random walk. 
Proposals belong to the same class should form a subgraph $\A \subset \X$. 
For feature aggregation, we are especially interested in minimizing the risk of incorrectly aggregating the features of a proposal which does not belong to the reference class. 
This risk can be measured by the transition probability $P_{\bar{\A}\A}$ from the subgraph $\bar{\A} = \X - \A$ to the subgraph $\A$.

The transition probability between subgraphs is formally defined as,
\begin{equation}
P_{\bar{\A}\A} = \dfrac{\sum_{i\in \bar{\A}, j\in \A}\pi_i \T_{ij}}{\sum_{i\in\bar{\A}}\pi_i},
\end{equation}
where $\pi_i = \sum_{k} \W_{jk} / \sum_{j,k} \W_{jk}$ denotes the stationary distribution of the graph. $\pi_i$ represents the connection strength between a proposal and the rest proposals in a graph.

As proved in ~\cite{meila2001random}, the transition probability is equivalent to the normalized minimum cut,
\begin{equation}
\text{NCut}(\A, \bar{\A}) = P_{\A\bar{\A}} + P_{\bar{\A}\A}.
\end{equation}

From the traditional spectral clustering view, the stochastic matrix $\T$ is fixed, and the transition probability is minimized by finding the optimal partition $\A, \bar{\A}$. 
However, from the supervised deep learning view, the stochastic matrix $\T$ derived from proposal features is the variable to optimize, and the optimal partition $\A, \bar{\A}$ is given. 
The optimization of $\T$ is further propagated to the proposal features and backbone network for discriminative feature learning. 
Furthermore, ~\cite{meila2001random} gives the desired form of $\T$, a blockwise diagonal matrix w.r.t $\A, \bar{\A}$, which is exactly the desired guide for proposal feature aggregation.

\subsection{Connection to Graph Convolution Network}
\label{sec:discuss}
Recently, Wang \etal~\cite{wang2018videos} have applied GCN for video classification task. 
They built a space-time graph with a similar affinity measurement to us. 
In their work, they took the edges of a graph as a general relation in space-time and mainly focus on modeling the high order interaction of objects in a video. 
However, in our work, we design the SELSA module to refine the features of a reference proposal by the relationship between them, which leads to a different motivation and optimization objective.

\section{Experiments on ImageNet VID}
In this section, we first introduce the datasets and evaluation metrics used for VID in Sec.~\ref{sec:dataset}, then followed by the implementation details of our method in Sec.~\ref{sec:impl}. 
We next justify the design choice of our SELSA module in Sec.~\ref{sec:abla} by ablation studies. 
We also investigate the effects of existing post-processing techniques on our method. 
Finally, we compare our method with other state-of-the-art methods.

\subsection{Dataset and Evaluation Setup}
\label{sec:dataset}
We train our model with a mixture of ImageNet VID and DET datasets with the split provided in FGFA~\cite{zhu17fgfa}.
We evaluate our proposed method on ImageNet VID dataset~\cite{ILSVRC15}. 
We report the mAP@IoU=0.5 and motion-specific mAP on the validation set. 

\subsection{Implementation Details}
\label{sec:impl}
\noindent\textbf{Feature Network} 
We use ResNet-101~\cite{he2016deep} as the backbone network for ablation studies. 
ResNeXt-101-$32\times 4$d~\cite{xie2017aggregated} is also used for the final results. 
The total stride of \emph{conv5} block is changed from 32 to 16 with dilated convolutions.

\noindent\textbf{Detection Network} 
RPN is applied on the output of \emph{conv4}. 
Anchors of 3 scales and 3 aspect ratios are used. 
Then Fast R-CNN is applied on the output of \emph{conv5}. 
We apply two fully connected (FC) layers upon the RoI pooled features followed by classification and bounding box regression. 

\noindent\textbf{SELSA Module}
We insert two SELSA modules into our network. 
Each one is inserted after one fully-connected layer in Faster R-CNN (FC $\rightarrow$ SELSA $\rightarrow$ FC $\rightarrow$ SELSA). 
The general transformation functions in Eq.~\ref{eq:similar} are instantiated as one fully-connected layer.

\noindent\textbf{Training and Testing Details}
The backbone networks are initialized with ImageNet pre-trained weights. 
A total of 220k iterations of SGD training is performed with a total batch size of 4 on 4 GPUs. 
The initial learning rate is $2.5 \times 10^{-4} $ and is divided by 10 at the 110k and the 165k iterations. 
For training, one training frame is sampled along with two random frames from the same video (identical frames for the DET dataset). 
For inference, $K$ frames from the same video are sampled along with the inference frame. 
In both training and inference, the images are resized to a shorter side of 600 pixels.

\subsection{Ablation study}
In this subsection, we study the impact of each design choice and parameter settings.
\label{sec:abla}
    \begin{table}[tb]
    \centering
        \begin{tabular}{l|c|c|c}
        \toprule
        Component & (a) & (b) & (c) \\ \midrule
        Semantics Aggregation &    & \checkmark & \checkmark\\ 
        Sequence-level Info &   & & \checkmark \\  \midrule
        mAP (\%)           & 73.62  & 75.26$_{\uparrow 1.64}$   & \textbf{80.25$_{\uparrow \textbf{6.63}}$} \\  \midrule
        mAP (\%) (slow)    & 82.12  & 83.59$_{\uparrow 1.47}$   & \textbf{86.91$_{\uparrow \textbf{4.79}}$} \\  \midrule
        mAP (\%) (medium)  & 70.96  & 72.88$_{\uparrow 1.92}$   & \textbf{78.94$_{\uparrow \textbf{7.98}}$} \\  \midrule
        mAP (\%) (fast)    & 51.53  & 51.43$_{\downarrow 0.10}$ & \textbf{61.38$_{\uparrow \textbf{9.85}}$} \\  \bottomrule
        \end{tabular}
        \caption{Detection results on the ImageNet VID validation set. For sequence-level methods, 21 frames are used when testing. No post-processing techniques are used. The absolute gains compared with the baseline are shown in the subscript.}
        \label{table:ablation_study}
    \end{table}

\begin{figure*}[htb]
	\centering
	\subfigure[]{
		\label{fig:abl:a}
		\includegraphics[scale=0.43]{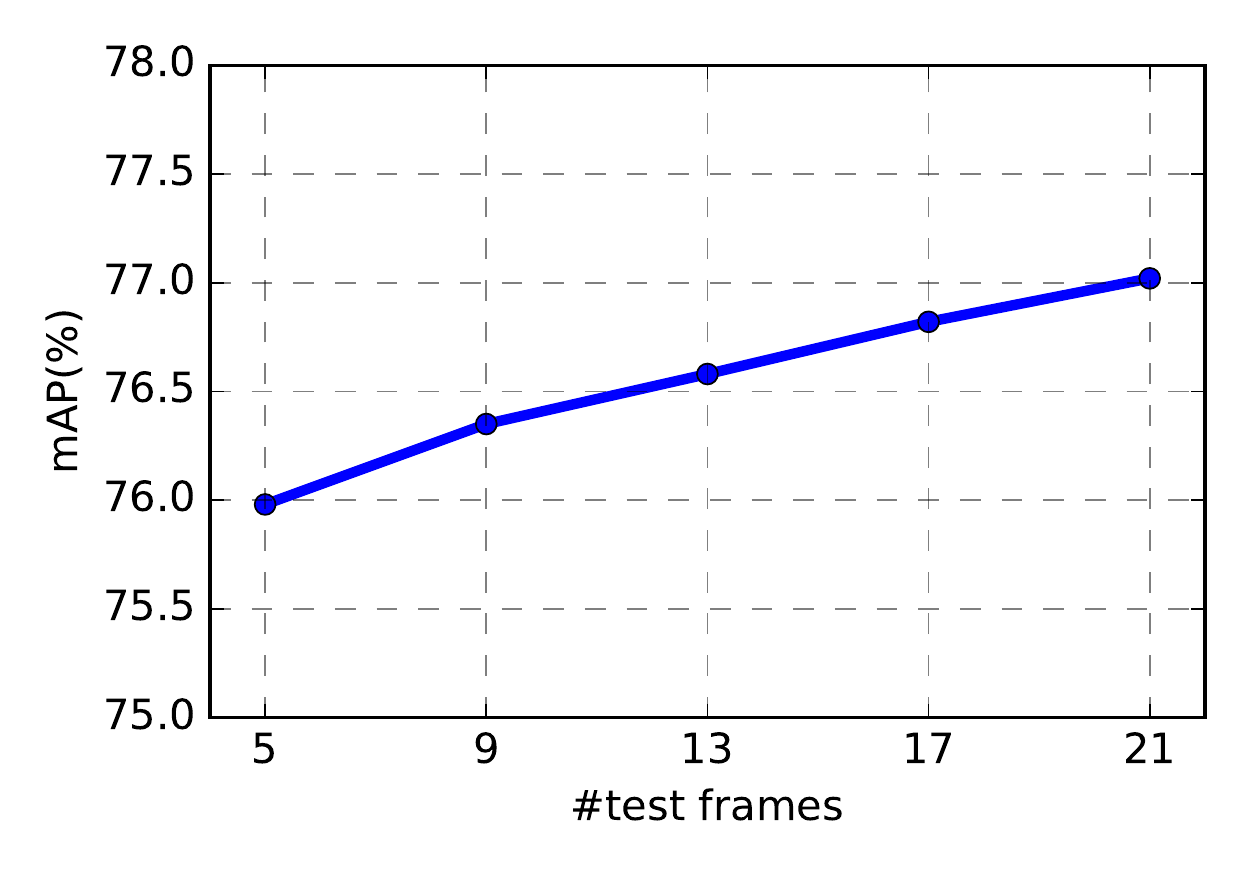}
	}
	\subfigure[]{
		\label{fig:abl:b}
		\includegraphics[scale=0.43]{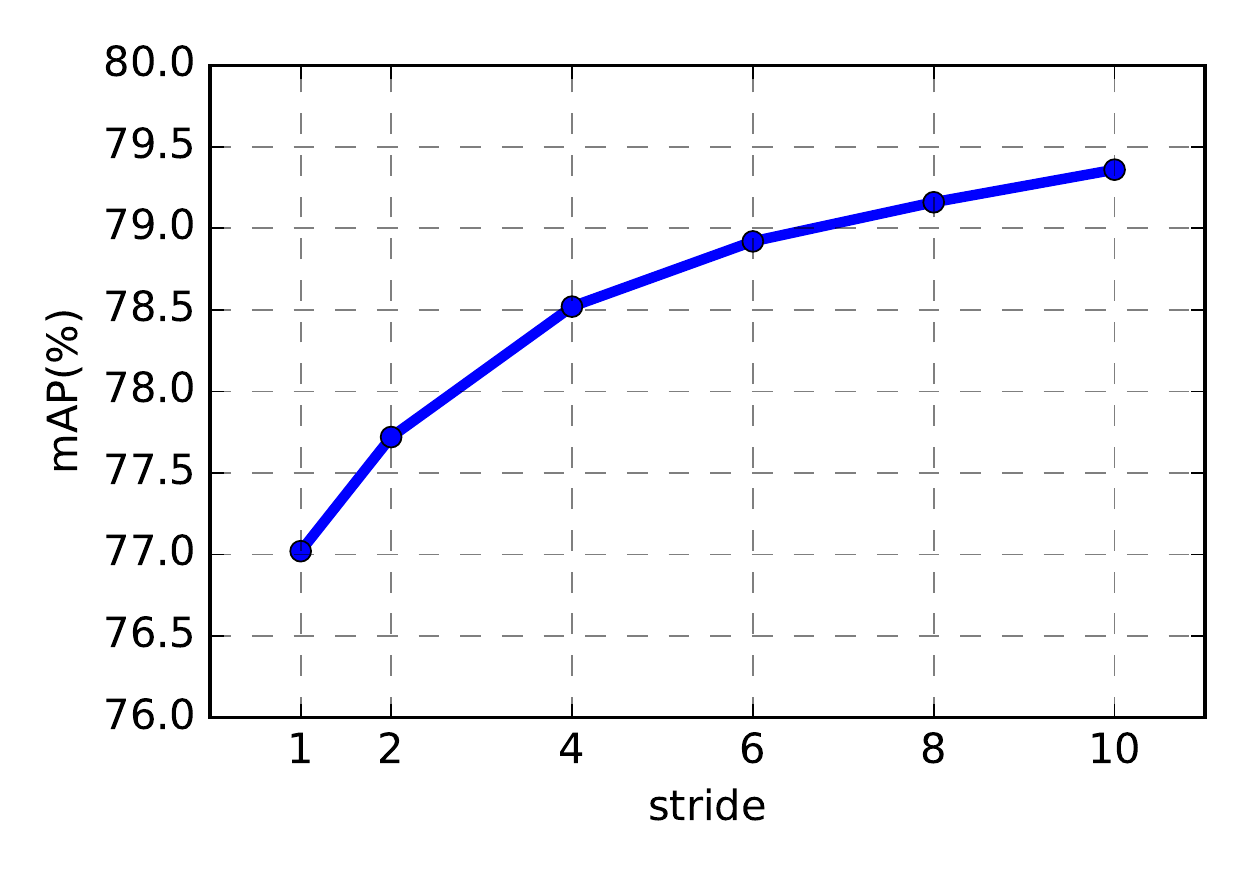}
	}
	\subfigure[]{
		\label{fig:abl:c}
		\includegraphics[scale=0.43]{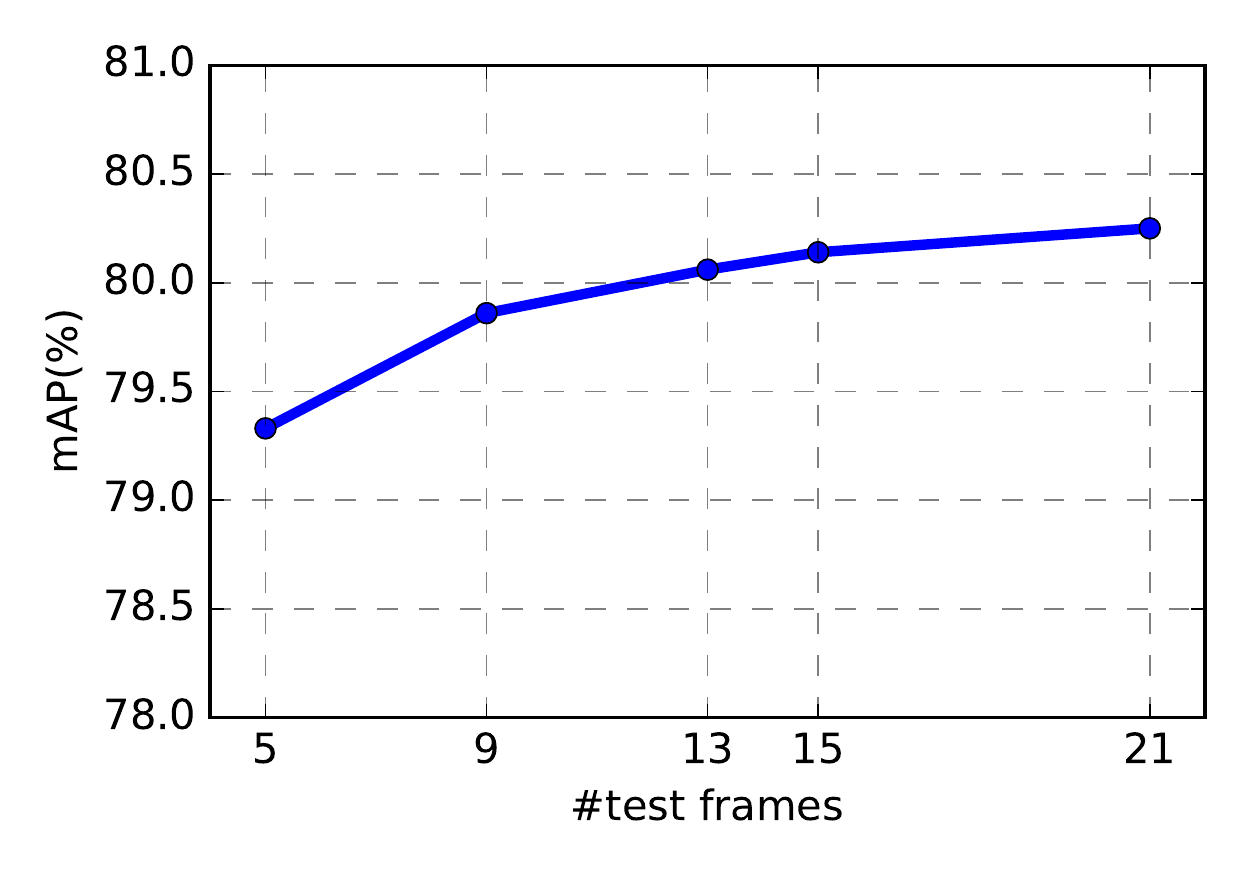}
	}
	\caption{Ablation analyses of different test settings. (a) The effect of different number of frames on sequential test performance. (b) The effect of different sampling stride on sequential test performance. (c) The effect of different number of frames on shuffled test performance. }
	\label{fig:sigma}
	\end{figure*}

\noindent
\paragraph{Effectiveness of SELSA} 
Table~\ref{table:ablation_study} compares our proposed methods with the single-frame baseline. 

Column (a) shows the results of our single-frame baseline. 
It uses ResNet-101 as the backbone and achieves a reasonable mAP of 73.62 as in~\cite{zhu17fgfa}. 

Column (b) performs semantics aggregation (SA) within a single frame, a degenerated variant of SELSA. 
More specifically, only proposals obtained from the same frame are considered as possible semantic neighbors for aggregation. 
This leads to a gain of 1.64 mAP compared with the baseline. 
When multiple objects with the same semantics or multiple proposals corresponding to the same object appear in the same frame, the semantically aggregated proposal features are hence enhanced with contextual information like in \cite{hu2018relation,chen2018context}, thus leading to the performance improvement. 
Note that for objects under fast motion, the mAP (fast) receives no improvement over baseline. 
This indicates that appearance degradation induced by fast motion could not be remedied by the contextual or object interaction information. 

Column (c) is the proposed SELSA method. 
It utilizes the SELSA module to enhance proposal features by sampling semantic neighbors from the full video sequence. 
It gives an mAP of 80.25, a large 6.63 mAP improvement compared with the baseline method. 
Note that it enhances the motion-specific performance in fast motion to 61.38 mAP, which is a huge improvement of 9.95 mAP compared with the baseline. 
Compared with column (b) and (c), it is easy to see that our method directly harvests high-quality features from aggregating sequence level features other than high order interaction information on the graph, as previously stated in Sec.~\ref{sec:discuss}.


\noindent
\paragraph{Sampling Strategies for Feature Aggregation}
Frame sampling strategy matters for video detection. 
As previous works~\cite{xiao2018video, zhu17fgfa} pointed out, using more frames in feature aggregation during testing yields better results. 
Besides, \cite{xiao2018video} samples frames with a uniform stride during testing to improve the performance. 

We examine the influence of the number of frames used and sampling strides when testing our method. 
Specifically, by using a sampling stride of $S$, one frame in every $S$ frames is used for testing instead of consecutive frames. 

First, we use sampling stride one and vary the number of frames used in aggregation.
As seen in Figure~\ref{fig:abl:a}, with more frames used for testing, the performance increases consistently. 
For example, using 21 frames for aggregation instead of 5 contributes a 1.04 mAP improvement. 

We then fix the number of frames for aggregation to 21 and examine the impact of sampling stride. 
Figure~\ref{fig:abl:b} shows the performance with different sample strides. 
Increasing the sampling stride from 1 to 10 further improves the performance from 77.02 to 79.36 mAP (a gain of 2.34 mAP). 
Notice that the sampling stride demonstrates a larger influence on the performance than the number of testing frames in general, which coincides with our assumption that our sequence level method could benefit more from sample diversity. 
Other feature aggregation methods which use optical flow or RNN may not benefit from the larger stride since it violates the temporal continuity assumption of these methods.

\noindent
\paragraph{Semantics Aggregation in Sequence Level} 
As discussed earlier, good features for aggregation in VID should be more diverse in terms of appearance and poses. This observation motivates the use of semantic neighbors instead of temporal neighbors.
Thus, taking a step further, we sample semantic neighbors uniformly from the full video sequence regardless of the temporal orders (shuffled test setting).
This is feasible since our method does not rely on any temporal information (e.g. optical flow), and also no feature alignment operation across frames is performed.
Our method is exempt from possible inaccurate predictions of temporal information (e.g. optical flow estimation~\cite{zhu17fgfa}, bounding box shifting prediction~\cite{feichtenhofer2017detect}) and feature alignment process~\cite{zhu2017deep, wang2018fully}, which is important when the motion is large. 
In fact, performance drops have been shown in optical flow based method~\cite{wang2018fully} as the number of frames increase when exceeding a certain threshold (12 frames in~\cite{wang2018fully}). 
Our method, on the contrary, shows its power of performing feature aggregation in the whole video sequence level in Figure~\ref{fig:abl:c}. 
As we have seen, using only 5 frames in shuffled test already achieves the same level of performance as 21 frames in strided testing. 
And using 21 frames along with shuffled testing gives an mAP of 80.25. 
This introduces an improvement of 0.89 mAP against to the strong result of 79.36 mAP where a sampling stride of 10 and in total 21 frames are used. 
This gain comes from sampling more diverse features in semantic neighbors rather than temporal neighbors, which further shows the effectiveness of SELSA for capturing the full sequence level information for feature aggregation. This is the default test setting in the following experiments.

\noindent 
\paragraph{Data augmentation} 
Existing VID datasets usually suffer from lacking of semantic diversity. 
Frames in a video are high similar to each other and thus lead to potential overfitting.
Thus we adopt data augmentation to alleviate this problem. Photometric distortion, random expand and random crop as in~\cite{liu2016ssd} are used besides the original random flipping operation. 
This gives us an improvement of 2.44 mAP, leading to 82.69 mAP when using ResNet-101 backbone.

\subsection{Video-level post-processing techniques}
\label{sec:comp_seq}
One advantage of our method is that it does not rely on post-processing methods (e.g Seq-NMS) to incorporate the full-sequence level information. 
Nearly all the state-of-the-art video detection systems~\cite{zhu17fgfa, wang2018fully, feichtenhofer2017detect, STLattice2018CVPR, xiao2018video} adopted post-processing methods which gives huge gains in performance. 
To illustrate that our method has already captured the full-sequence level information, we further apply the Seq-NMS post-processing upon our method. 
Table~\ref{table:seq_nms} shows the results of how Seq-NMS affects our methods when using different backbone networks.
As easily seen, adding Seq-NMS only has a minor impact on the results. 
In particular, adding Seq-NMS to ResNet-101/ResNext-101 backbone network yields 0.21/0.57 mAP drop. 

Referring to Table~\ref{table:compare_stoa}, post processing methods have introduced large performance improvement upon existing state-of-the-art methods: 2.1 mAP for FGFA~\cite{zhu17fgfa} and 2.2 mAP for MANet~\cite{wang2018fully} with Seq-NMS and 4 mAP for D (\& T loss)~\cite{feichtenhofer2017detect} with tubelet rescore. 
In contrast, almost no gain from Seq-NMS upon our method with ResNet-101 as backbone network shows that our method has already largely captured the full-video level information through our SELSA module without any post-processing techniques. 
Moreover, different from post-processing methods like Seq-NMS which involves two separate stages, our method could be trained end-to-end with sequence level information. 
As the backbone feature network becomes stronger, our method could even better utilize such sequence level information, thus shows a better result than that with Seq-NMS, in which the separate post-processing steps might lead to sub-optimal results.

\begin{table}[tb]
    \centering
        \begin{tabular}{c|c|c|c|c}
        \toprule
        Backbone & \multicolumn{2}{c|}{ResNet-101} & \multicolumn{2}{c}{ResNeXt-101} \\ 
        \midrule
        Seq-NMS  &     &    \checkmark &                 &  \checkmark \\ \hline
        mAP (\%) &     82.69           &  82.48  $_{\downarrow 0.21}$&    84.30&    83.73$_{\downarrow 0.57}$\\ 
        \bottomrule
        \end{tabular}
        \caption{The effects of post-processing on our method. The absolute gains compared with the method without Seq-NMS are shown in the subscript.}
        \label{table:seq_nms}
\end{table}

\subsection{Comparison with the state-of-the-art methods}
    \begin{table}[htb]
        \centering
        \begin{tabular}{l|c|c}
        	\toprule
             Methods & Backbone &  mAP (\%) \\ \midrule \midrule
             FGFA~\cite{zhu17fgfa} & \multirow{4}{*}{ResNet-101} &  76.3 \\ 
             D (\& T loss)~\cite{feichtenhofer2017detect} &  &  75.8 \\
             MANet~\cite{wang2018fully} &  &  78.1 \\
             Ours & &   \textbf{80.25}\\
             \midrule
             FGFA*~\cite{zhu17fgfa} & \multirow{6}{*}{ResNet-101} &  78.4 \\ 
             MANet*~\cite{wang2018fully} &  &  80.3 \\
             ST-Lattice*~\cite{STLattice2018CVPR} &  &  79.6 \\
             D\&T*~\cite{feichtenhofer2017detect} &  & 79.8 \\
             STMN*+~\cite{xiao2018video} &  &  80.5 \\
             Ours* &  &   \textbf{80.54}\\
             Ours$^\triangle$ &  &   \textbf{82.69}\\
             \midrule
             D\&T*~\cite{feichtenhofer2017detect} & ResNeXt-101 &  81.6 \\
             D\&T*~\cite{feichtenhofer2017detect} & Inception-v4 &  82.1 \\
             Ours & ResNeXt-101 &   \textbf{83.11} \\
             Ours$^\triangle$ & ResNeXt-101 &   \textbf{84.30} \\
             \bottomrule
        \end{tabular}
        \caption{Performance comparison with state-of-the-art systems on the ImageNet VID validation set. * indicates use of video-level post-processing methods (e.g Seq-NMS, tubelet rescoring). + indicates use of model emsembling. $^\triangle$ indicates using data augmentation.}
        \label{table:compare_stoa}
    \end{table}


\begin{figure*}[!ht]
    	\centering
    	\includegraphics[width=1.0\textwidth]{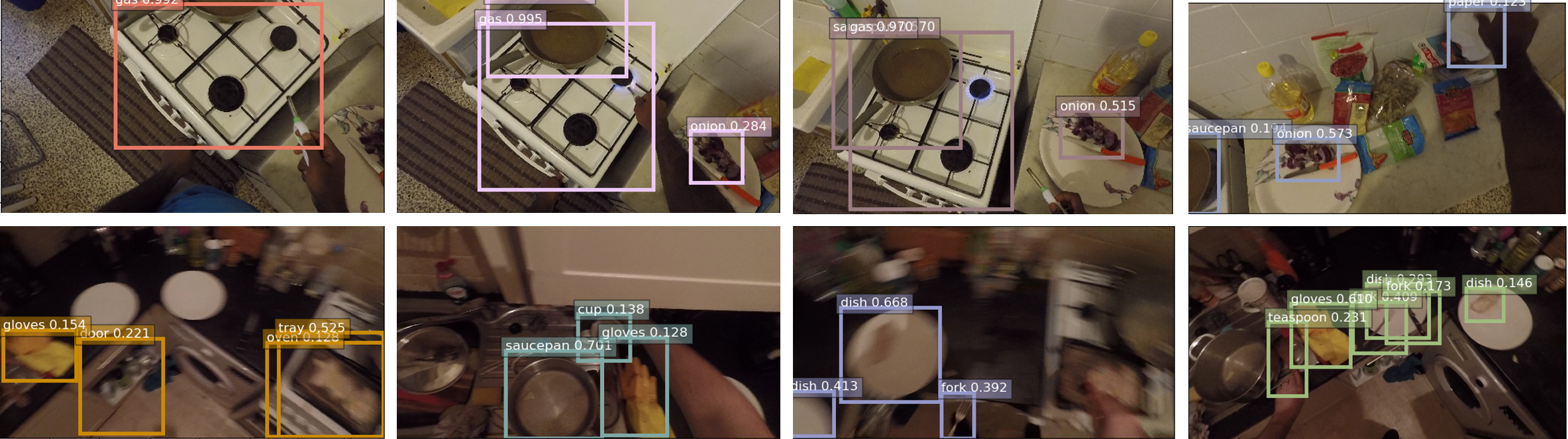}
     	\vspace{2mm}
        \caption{Visual results of our method on EPIC KITCHENS.}
    	\label{fig:epic_results}
\end{figure*}

Table~\ref{table:compare_stoa} summarizes the performance of our methods and other state-of-the-art methods on the ImageNet VID validation set. Our method achieves the best performance among various testing settings. 

With no video-level post-processing techniques, compared with FGFA~\cite{zhu17fgfa} (76.3 mAP) and MANet~\cite{wang2018fully} (78.1 mAP) which are both built on flow-based feature aggregation, our method is remarkably better (80.25 mAP), outperforming these two methods by 3.95 and 2.15 mAP, respectively. 
It also outperforms D (\& T loss)~\cite{feichtenhofer2017detect} by a large margin of 4.45 mAP.

The middle part of Table~\ref{table:compare_stoa} shows the comparison with methods that utilize sequence-level post-processing techniques. 
FGFA*, MANet* and STMN*+~\cite{xiao2018video} use Seq-NMS, while D\&T*~\cite{feichtenhofer2017detect}, ST-Lattice*~\cite{STLattice2018CVPR} utilize tubelet rescoring. 
Our method, by using Seq-NMS as the post-processing method, achieves 80.54 mAP, which is slightly better than the previous state-of-the-art method STMN*+. 

Furthermore, by plugging in the stornger ResNeXt-101, our method achieves performance of 83.11 mAP \textbf{without} any post-processing techniques (e.g Seq-NMS),  which surpasses the D\&T with the same backbone and tubulet rescoring by a large margin (1.15 mAP). 
Our method benefits from the stronger representation power introduced by better backbone networks. 
When equipped with training data augmentation, our methods show a significant gain of 2.44/1.19 mAP for ResNet-101/ResNeXt-101. This indicates SELSA can benefit from the diversity of proposal features during aggregation.
These results reveal the potential of our proposed method.

\section{Additional Experiments on Epic Kitchen}
ImageNet VID dataset falls short in the density and diversity of objects. Here we evaluate SELSA on the EPIC KITCHENS dataset~\cite{damen2018scaling}.

\subsection{Dataset and Evaluation Setup}
EPIC KITCHENS~\cite{damen2018scaling} is a large scale egocentric dataset, capturing daily activities happened in the kitchens. In EPIC KITCHENS dataset, each frame contains avg/max 1.7/9 objects, which is far more complex and challenging. The video object detection task consists of 32 different kitchens with 454,255 object bounding boxes spanning 290 classes. 272 video sequences captured in 28 kitchens are used for training. 106 sequences collected in the same 28 kitchens (S1) and 54 sequences collected in other 4 unseen kitchens (S2) are used for evaluation. Videos are annotated in 1s interval.

\subsection{Implementation Details}
Mostly, we adopt the same network setting as on ImageNet VID dataset. 
No data augmentation except random horizontal flip is used. 
A total of 600k iterations of SGD training is performed on 4 GPUs. 
The initial learning rate is $2.5 \times 10^{-4} $ and is divided by 10 at the 300k iterations. 
For both training and inference, we sample frames within a $\pm 10$s window for the SELSA module.

\subsection{Results and Analysis}

\begin{table}[htb]
\centering
\begin{tabular}{l|c|c|c}
\hline 
& \multicolumn{3}{c}{S1} \\ 
\hline
Methods & mAP@.05 & mAP@.5 & mAP@.75 \\  
\hline
EPIC~\cite{damen2018scaling}        & 45.99                 & 34.18                & 8.49                   \\ 
\hline
Faster R-CNN & 53.12                 & 36.57                & 9.97                   \\ 
\hline
Ours & 54.67                 & \textbf{37.97}                & 9.81                   \\ 
\hline
\midrule \midrule
& \multicolumn{3}{c}{S2} \\ 
\hline
Methods & mAP@.05          & mAP@.5          & mAP@.75          \\  
\hline
EPIC~\cite{damen2018scaling} & 44.95                 & 32.01                & 7.87                   \\ 
\hline
Faster R-CNN& 48.91                 & 31.86                & 7.36                   \\  
\hline
Ours &  50.25                & \textbf{34.80}                & 8.10                   \\ 
\hline
\end{tabular}
\caption{Performance comparison on EPIC KITCHENS test set. S1 and S2 indicate Seen and Unseen splits.}
\label{tb:epic}
\end{table}

Here we present some preliminary results on the EPIC KITCHENS dataset. 
As shown in Table~\ref{tb:epic}, SELSA improves over Faster R-CNN baseline by 1.4/2.94 mAP for Seen/Unseen splits.  
Although the training scheme and the hyper parameter selection are far from optimal, 
our method still achieves promising results. 
This shows that SELSA is applicable to more complex video detection tasks. Figure~\ref{fig:epic_results} shows some results of our method.

\section{Conclusion}
\normalsize
In this work, we have proposed a novel view of VID problem by taking the full-sequence level feature aggregation. 
Instead of using methods such as optical flow or RNN, we propose a simple yet effective SELSA module for aggregating semantic features across frames. 
Since the aggregation is conducted on the proposal level rather than feature map or even pixel level, our method is more robust to motion blur and large pose variation. 
Furthermore, we have derived the connection between our method and the classic spectral clustering method, providing a novel clustering view of our method. 
Extensive ablation analyses demonstrate the effectiveness of the proposed SELSA module. 
When compared with previous methods, our method achieves superior performance without sophisticated post-processing methods.

\section*{Acknowledge}
This work was supported in part by the National Key R\&D Program of China (No. 2018YFB1402605), the Beijing Municipal Natural Science Foundation (No. Z181100008918010), the National Natural Science Foundation of China (No. 61836014, No. 61761146004, No. 61773375, No. 61602481). The authors would like to thanks NVAIL for the support.

{
\small
\bibliographystyle{ieee_fullname}
\bibliography{egbib}
}

\end{document}